\documentclass[11pt,a4paper]{article}
\usepackage[hyperref]{acl2021}
\usepackage{times}
\usepackage{latexsym}

\usepackage{graphicx}
\usepackage{amsmath}
\usepackage{booktabs}
\usepackage{multirow}
\usepackage{microtype}
\usepackage{cancel}
\usepackage{soul}
\usepackage{url}
\usepackage{latexsym}
\usepackage{bm}
\usepackage{helvet}
\usepackage{courier}
\usepackage{caption}
\usepackage{hyperref}
\usepackage{amsfonts}
\usepackage{enumitem}

\aclfinalcopy % Uncomment this line for the final submission
\title{Element Intervention for Open Relation Extraction}

  \author{Fangchao Liu${}^{1,3}$, Lingyong Yan${}^{1,3}$, Hongyu Lin${}^{1,}$\thanks{~ Corresponding authors.}\ \  , Xianpei Han${}^{1,2,*}$, Le Sun${}^{1,2}$ \\
  ${}^{1}$Chinese Information Processing Laboratory ~ ${}^{2}$State Key Laboratory of Computer Science \\
  Institute of Software, Chinese Academy of Sciences, Beijing, China\\
  ${}^{3}$University of Chinese Academy of Sciences, Beijing, China \\
   {\tt \{fangchao2017,lingyong2014,hongyu,xianpei,sunle\}@iscas.ac.cn}
  }

\date{}

\begin{document}
\maketitle

\begin{abstract}
  Open relation extraction aims to cluster relation instances referring to the same underlying relation, which is a critical step for general relation extraction. Current OpenRE models are commonly trained on the datasets generated from distant supervision, which often results in instability and makes the model easily collapsed.  In this paper, we revisit the procedure of OpenRE from a causal view. By formulating OpenRE using a structural causal model, we identify that the above-mentioned problems stem from the spurious correlations from entities and context to the relation type. To address this issue, we conduct \emph{Element Intervention}, which intervenes on the context and entities respectively to obtain the underlying causal effects of them. We also provide two specific implementations of the interventions based on entity ranking and context contrasting. Experimental results on unsupervised relation extraction datasets show that our methods outperform previous state-of-the-art methods and are robust across different datasets\footnote{Code available at \href{https://github.com/Lfc1993/EI\_ORE}{https://github.com/Lfc1993/EI\_ORE}}.
\end{abstract}

\section{Introduction}
Relation extraction~(RE) is the task to extract relation between entity pair in plain text. 
For example, when given the entity pair \textit{(Obama, the United States)} in the sentence \textit{``Obama was sworn in as the 44th president of the United States"}, an RE model should accurately predict the relationship \textit{``President\_of"} and extract the corresponding triplet \textit{(Obama, President\_of, the United States)} for downstream tasks.
Despite the success of many RE models~\citep{zeng-2014, soares-2019}, most previous RE paradigms rely on the pre-defined relation types, which are always unavailable in open domain scenario and thereby limits their capability in real applications.

\begin{figure}[!t] 
  \setlength{\belowcaptionskip}{-0.4cm}
  \centering
  \includegraphics[width=0.45\textwidth]{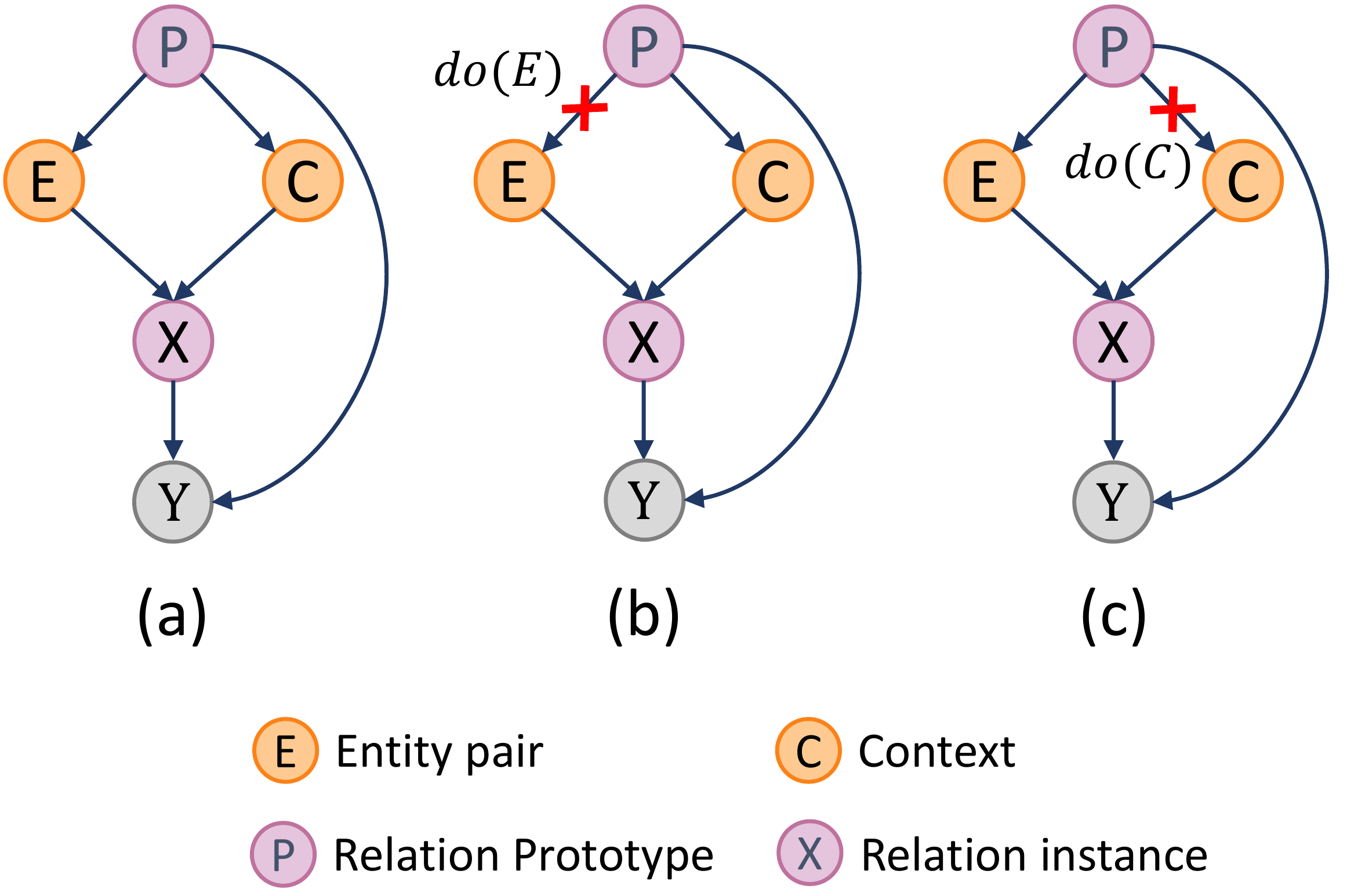}
    \caption{The Structural Causal Model demonstrates the procedure of OpenRE. (a) is the original SCM; (b) Entity intervention that fixes the entity pair and adjusts different contexts; (c) Context intervention that fixes the context and adjusts different entity pairs.}
  \label{fig:scm}
\end{figure}

Open Relation Extraction~(OpenRE), on the other hand, has been proposed to extract relation facts without pre-defined relation types neither annotated data. Given a relation instance consisting of two entities and their context, OpenRE aims to identify other instances which mention the same relation. To achieve this, OpenRE is commonly formulated as a clustering or pair-matching task. Therefore the most critical challenge for OpenRE is how to learn effective representations for relation instances and then cluster them. To this end, ~\citet{yao-2011} adopt topic model~\citep{blei-lda} to generate latent relation type for unlabelled instances. Later works start to utilize datasets collected using distant supervision for model training. Along this line, \citet{marcheggiani-2016} utilize an auto-encoder model and trains the model through self-supervised signals from entity link predictor. \citet{hu-2020-selfore} encode each instance with pre-trained language model~\citep{devlin-2019,soares-2019} and learn the representation by self-supervised signals from pseudo labels.

Unfortunately, current OpenRE models are often unstable and easily collapsed~\cite{simon-2019}. For example, OpenRE models frequently cluster all relation instances with context ``was born in'' into the relation type \emph{BORN\_IN\_PLACE} because they share similar context information. However, ``was born in'' can also refer to the relation \emph{BORN\_IN\_TIME}. Furthermore, current models also tend to cluster two relation instances with the same entities (i.e., relation instances with the same head and tail entities) or the same entity types into one relation. This problem can be even more severe if the dataset is generated using distant supervision because it severely relies on prototypical context and entity information as supervision signals and therefore lacks of diversity.

In this paper, we attempt to explain and resolve the above-mentioned problem in OpenRE from a causal view. Specifically, we formulate the process of OpenRE using a structural causal model~(SCM)~\citep{pearl_2009}, as shown in Figure~\ref{fig:scm}. The main assumption behind the SCM is that distant supervision will generate highly correlated relation instances to the original prototypical instance, and there is a strong connection between the generated instance to the prototypical instance through either their entities or their context. For example, ''[Jobs] was born in [California]'' and ''[Jobs] was born in [1955]'' are highly correlated because they share similar context ``was born in'' and entity ``Jobs''. Such connection will result in spurious correlations, which appear in the form of the backdoor paths in the SCM. Then the spurious correlations will mislead OpenRE models, which are trained to capture the connection between entities and context to the relation type.

Based on the above observations, we propose \emph{element intervention}, which conducts backdoor adjustment on entities and context respectively to block the backdoor paths. However, due to the lack of supervision signals, we cannot directly optimize towards the underlying causal effects. To this end, we further propose two surrogate implementations on the adjustments on context and entities, respectively. Specifically, we regard the instances in the original datasets as the relation prototypes. Then we implement the adjustment on context through a Hierarchy-Based Entity Ranking~(Hyber), which fixes the context, samples related entities from an entity hierarchy tree and learns the causal relation through rank-based learning. Besides, we implement the adjustment on entities through a Generation-based Context Contrasting~(Gcc), which fixes the entities, generates positive and negative contexts from a generation-based model and learns the causal effects through contrastive learning.

We conduct experiments on different unsupervised relation extraction datasets. Experimental results show that our methods outperform previous state-of-the-art methods with a large margin and suffer much less performance discrepancy between different datasets, which demonstrate the effectiveness and robustness of the proposed methods.

\begin{figure*}[!ht]
  \setlength{\belowcaptionskip}{-0.4cm}
\centering
\includegraphics[width=0.99\textwidth]{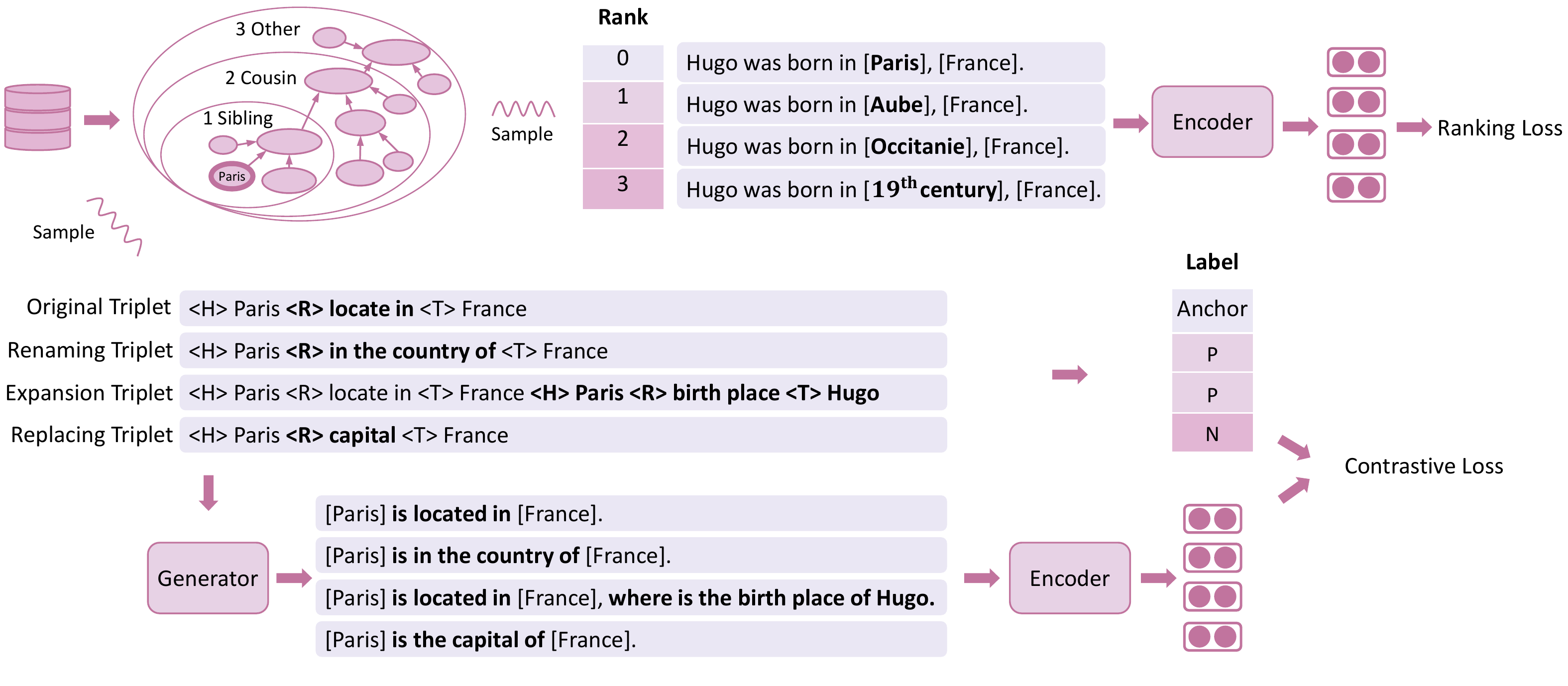}
  \caption{Framework of Element Intervention.
  }
\label{fig:framework}
\end{figure*}

\section{OpenRE from Causal View}
In this section, we formulate OpenRE from the perspective of Structural Causal Model and give the theoretical proof for intervention methods that block the backdoor paths from relation elements~(i.e., context and entity pair) to the latent relation types.

\subsection{Task Definition}
Relation extraction~(RE) is the task of extracting the relationship between two given entities in the context.
Considering the sequence example: $\mathcal{S}=[s_0,...,s_{n-1}]$ which contains $n$ words,
$e_1=[i,j]$ and $e_2=[k,l]$ indicate the entity pair, where $0\le i \le j < k\le l \le n-1$,
a relation instance $X$ is defined as $X = (\mathcal{S},e_1,e_2)$, (i.e. the tuple of entity pair and the corresponding context).
The element of a relation instance is the entity pair and the corresponding context.
Traditional RE task is to predict the relations type when given $X$.
However, the target relation types are not pre-defined in OpenRE. Consequently, OpenRE is commonly formulated as a clustering task or a pair-matching task by considering whether two relation instances $X_i$ and $X_j$ refer to the same relation.

Unfortunately, current OpenRE models are often unstable and easily collapsed~\cite{simon-2019}. In the next section, we formulate OpenRE using a structural causal model and then identify the reasons behind these deficiencies from the SCM.

\subsection{Structural Causal Model for OpenRE}
Figure~\ref{fig:scm}~(a) shows the structural causal model for OpenRE. The main idea behind the SCM is distant supervision will generate highly correlated relation instances to the original prototypical instance, and there is a strong connection between the generated instance to the prototypical instance through either their entities or their context. Specifically, in the SCM, we describe OpenRE with five critical variables: 1) the prototypical relation instance $P$, which is a representative relation instance of one relation type cluster; 2) the entity pair  $E$, which encodes the entity information of one relation instance; 3) the context $C$, which encodes the context information of one relation instance; 4) a relation instance $X$ (which can be generated from distant supervision or other strategies) and 5) the final pair-wise matching result $Y$, which corresponds to whether instance $X$ and the prototypical relation instance $P$ entail the same relation.

Given the variables mentioned above, we formulate the process of generating OpenRE instances based on the following causal relations:\begin{itemize}
  \item $E\leftarrow P\rightarrow C$ formulates the process of sampling related entities and context respectively from the prototypical relation instance $P$.
  \item $E\rightarrow X\leftarrow C$ formulates the relation instance generating process. Given the context $C$ and entities $E$ from the prototypical relation instance $P$, a new relation instance $X$ is generated based on the information in C and E. This process can be conducted through distant supervision.
  \item $P \rightarrow Y\leftarrow X$ formulates the OpenRE clustering or pair-wise matching process. Given a prototypical relation instance $P$ and another relation instance $X$, this process will determine whether $X$ belongs to the relation cluster of $P$.
\end{itemize}

\subsection{Spurious Correlations in OpenRE}
Given a relation prototypical instance $P$, the learning process of OpenRE is commonly to maximize the probability $\mathcal{P}(y,P|X) = \mathcal{P}(y,P|E,C)$. However, as it can be observed from the SCM, there exists a backdoor path  $P\rightarrow E \rightarrow X$ when we learn the underlying effects of context $C$. That is to say, the learned effect of $C$ to $Y$ is confounded by $E$ (through $P$). For example, when we learned the effects of context ``was born in'' to the relation ``BORN\_IN\_PLACE'', the backdoor path will lead the model to mistake the contribution of the entities (PERSON, PLACE) to the contribution of context, and therefore resulted in spurious correlation. The same thing happens when we learn the effects of entities $E$, which is influenced by the backdoor path $P\rightarrow C \rightarrow X$. As a result, optimizing these spurious correlations will result in an unstable and collapsed OpenRE model.

\subsection{Resolving Spurious Correlations via Element Intervention}
To resolve the spurious correlations, we adopt the backdoor adjustment~\citep{pearl_2009} to block the backdoor paths.
Specifically, we separately intervene on context $C$ and entities $E$ by applying the \textit{do}--operation.

\paragraph{Entity Intervention.}
As shown in Figure~\ref{fig:scm}~(b), to avoid the spurious correlations of entities to relation types, we conduct the $do$-operation by intervening on the entities $E$:
\begin{equation}
  \setlength{\abovedisplayskip}{3pt}
  \setlength{\belowdisplayskip}{3pt}
  \begin{aligned}
    \mathcal{P}(Y,P|&do(E=e_0))\\
    % =&\sum_{c}\sum_{d}P(d,c)P(y|e_0,c)\\
    % =&\sum_{c}P(y|e_0,c)\sum_{d}P(d,c)\\
    =&\sum_{C,X}\mathcal{P}(C,P)\mathcal{P}(X,Y|e_0,C,P)\\
    =&\sum_{C}\mathcal{P}(C,P)\mathcal{P}(Y|e_0,C,P)\\
    =&\sum_{C}\mathcal{P}(P)\mathcal{P}(C|P)\mathcal{P}(Y|e_0,C,P)\\
  \end{aligned}
      \label{eq:ei_form}
\end{equation}
Since $\mathcal{P}(P)$ is uniformly distributed in the real world, this equation can be rewritten as:
\begin{equation}
  \setlength{\abovedisplayskip}{3pt}
  \setlength{\belowdisplayskip}{3pt}
  \begin{aligned}
    \mathcal{P}(Y,P|&do(E=e_0))\\
    \propto &\sum_{C}\mathcal{P}(C|P)\mathcal{P}(Y|e_0,C,P)
  \end{aligned}
    \label{eq:ei}
\end{equation}
% In this work, we assumpt the condition probability of $d$ when given $e$ is uniformly distributted, so Equation~\ref{eq:ci_former} can be simplified as:
% \begin{equation}
%   P(Y_U|do(C=c_0))=\sum_{e}P(e)P(Y_U|c_0,e)
%   \label{eq:ci}
% \end{equation}
This equation means the causal effect from the entities $E$ to its matching result $Y$ can be estimated by considering the corresponding possibility of each context given the prototypical relation instance $P$. The detailed implementation will be described in the next section.

\paragraph{Context Intervention.}
Similarly, we conduct context intervention to avoid the spurious correlations of context to relation types, as shown in Figure~\ref{fig:scm}~(c):
\begin{equation}
  \setlength{\abovedisplayskip}{3pt}
  \setlength{\belowdisplayskip}{3pt}
  \begin{aligned}
    \mathcal{P}(Y,P|&do(C=c_0))\\
    \propto &\sum_{E}\mathcal{P}(E|P)\mathcal{P}(Y|c_0,E,P)
  \end{aligned}
    \label{eq:ci}
\end{equation}
which means the causal effect from the context $C$ to its matching result $Y$ can be estimated by considering the corresponding possibility of each entity $E$ given $P$. The detailed implementation will also be described in the next section.

\subsection{Optimizing Causal Effects for OpenRE}
To effectively capture the causal effects of entities $E$ and context $C$ to OpenRE, a matching model $\mathcal{P}(Y|C,E,P;\theta)$ should be learned by optimizing the causal effects:
\begin{equation}
  \small
  \begin{aligned}
    L(\theta) = &I(X,P)\cdot \mathcal{P}(Y=1,P| do(E= e(X)) \\
     &+I(X,P)\cdot \mathcal{P}(Y=1,P| do(C= c(X)) \\
     &+[1-I(X,P)]\cdot \mathcal{P}(Y=0,P| do(E= e(X)) \\
     &+[1-I(X,P)]\cdot \mathcal{P}(Y=0,P| do(C= c(X)) \\
    \end{aligned}
    \label{eq:loss}
  \end{equation}
  where $e(X)$ and $c(X)$ represents the entities and context in relation instance $X$, $I(X,P)$ is an indicator which represents whether $X$ and $P$ belong to the same relation. $\mathcal{P}(Y|C,E,P;\theta) = \mathcal{P}(Y|X,P;\theta) $ is a matching model, which is defined using a prototype-based measurements:
\begin{equation}
  \begin{aligned}
    \mathcal{P}(Y|X,P;\theta) \propto - D(R(X;\theta),R(P;\theta))
  \end{aligned}
    \label{eq:distance}
\end{equation}
where $D$ is a distance measurement and $R(X;\theta)$ is a representation learning model parametrized by $\theta$, which needs to be optimized during learning. In the following, we will use $D(X,P) = D(R(X;\theta),R(P;\theta))$ for short.

However, it is difficult to directly optimize the above loss function because 1) in unsupervised OpenRE, we are unable to know whether the relation instance $X$ generated from $(E,C)$ matches the prototypical relation instance $P$; 2) we are unable to traverse all possible $E$ and $C$ in Equation (\ref{eq:ei}) and (\ref{eq:ci}). To resolve these problems, in the next section, we will describe how we implement the context intervention via hierarchy-based entity ranking and the entity intervention via generation-based context contrasting.

% Since the true label is unavailable in OpenRE, we adopt the proof from \citet{}, and use the contrastive~(Ranking)-based representation learning for intervention.

% \begin{figure}[!t] 
%   \setlength{\belowcaptionskip}{-0.4cm}
%   \centering
%   \includegraphics[width=0.45\textwidth]{figures/entity_tree.pdf}
%     \caption{Hierarchical Sibling Tree constructeed from the meta-information of WikiData. \textbf{"Paris"} is the original entity in the context, \textit{"Depart.", "Admini. Div."} is short for \textit{"Department" and "Administrative Division"}. In this work, we alternate the entity according to the level of similarity in the tree.}
%   \label{fig:et}
% \end{figure}

\section{Element Intervention Implementation}
As we mentioned above, it is difficult to directly optimize the causal effects via Equation (\ref{eq:loss}). To tackle this issue, this section provides a detailed implementation to approximate the causal effects. Specifically, we regard all relation instances in the original data as the prototypical relation instance $P$, and then generate highly correlated relation instances $X$ from $P$ via a hierarchy-based sampling and generation-based contrasting. Then we regard structural signals from the entity hierarchy and confidence score from the generator as distant supervision signals, and learn the causal effects via ranking-based learning and contrastive learning.

\subsection{Hierarchy-based Entity Ranking for Context Intervention}
To implement context intervention, we propose to formulate  $\mathcal{P}(E|P)$ using an entity hierarchy, and approximately learn to optimize the causal effects of $\mathcal{P}(Y=1,P| do(C))$ and $\mathcal{P}(Y=0,P| do(C))$ in Equation (\ref{eq:loss}) via a hierarchy-based entity ranking loss. Specifically, we first regard all relation instances in the data as prototypical relation instance $P$. Then we formulate the distribution $\mathcal{P}(E|P)$ by fixing the context in $P$ and replacing entities by sampling from an entity hierarchy. Each sampled entity is regarded as the same $\mathcal{P}(E|P)$.
Intuitively, the entity closer to the original entities in $P$ tends to generate more consistent relation instance to $P$. To approximate this semantic similarity, we utilize the meta-information in WikiData~(i.e., the \textit{``instance\_of"} and \textit{``subclass\_of"} statements, which describe the basic property and concept of each entity), and construct a hierarchical entity tree for ranking the similarity between entities.
In this work, we apply a three-level hierarchy through these two statements:
\begin{itemize}[leftmargin=*]
  \setlength{\itemsep}{1pt}
  \item \textbf{Sibling Entities}: The entities belonging to the same parent category as the original entity. For example, \textit{``Aube"} and \textit{``Paris"} are sibling entities since they are both the child entity of \textit{``department of France"}, and both express the concepts of location and GPE. These sibling entities can be considered as golden entities to replace.
  \item \textbf{Cousin Entities}: The entities belonging to the same grandparent category but the different parent category from the original entity. For example, \textit{``Occitanie"} and \textit{``Paris"} is of the same grandparent category~\textit{``French Administrative Division"}, but shares different parent category. These entities can be considered as silver entities since they are likely to be the same type as the original one but less possible than the sibling entities.
  \item \textbf{Other Entities}: The entities beyond the grandparent category, which are much less likely to be the same type as the original one.
\end{itemize}

For the example in Figure~\ref{fig:framework}, the prototypical relation instance \textit{``Hugo was born in [Paris], [France]"} is sampled to be intervened.
We first fix the context and randomly choose one of the head or tail entities to be replaced. In this case, we choose ~\textit{"Paris"}.
Then, entities that correspond to different hierarchies are sampled and to replace the original entity.
In this case, \textit{``Aube"} is sampled as the sibling entity, \textit{``Occitanie"} to be the cousin entity and \textit{``$19^{th}$ century"} to be the other entity.

After sampled these intervened instances, we approximately optimize $\mathcal{P}(Y,P| do(C)) $ using a rank-based loss function:
\begin{equation}
  \setlength{\abovedisplayskip}{3pt}
  \setlength{\belowdisplayskip}{3pt}
  \begin{aligned}
    \mathcal{L}_{E}(\theta;\mathcal{X})=\sum_{i=1}^{n-1}&\text{max}(0,\\
    &D(P,X_i)-D(P,X_{i+1})+m_E),
  \end{aligned}
\end{equation}
where $\theta$ is the model parameters, $D(X_i,P)$ is the distances between representations of generated relation instance $X_i$ and prototypical relation instance $P$. $X$ is the intervened relation instance set, $m_E$ is the margin for entity ranking loss, and $n=3$ is the depth of the entity hierarchy.

\subsection{Generation-based Context Contrasting for Entity Intervention}

Different from the context intervention that can easily replace entities, it is more difficult to intervene on entities and modify the context. Fortunately, the rapid progress in pre-trained language model~\citep{gpt2, bart,raffel-2020} makes the language generation from RDF data\footnote{\href{https://www.w3.org/TR/WD-rdf-syntax-971002/}{https://www.w3.org/TR/WD-rdf-syntax-971002/}} available~\citep{ribeiro2020}.
So in this work, we take a different paradigm named Generation-based Context Contrasting,  which directly generates different relation instances from specifically designed relation triplets, and approximately learn to optimize the causal effects of $\mathcal{P}(Y=1,P| do(E))$ and $\mathcal{P}(Y=0,P| do(E))$ in Equation (\ref{eq:loss}) via contrastive learning.
Specifically, we first sample relation triplets from Wikidata as prototypical relation instance $P$, and then generates relation triplets with the same entities but different relation context using the following strategies:
\begin{itemize}[leftmargin=*]
  \setlength{\itemsep}{1pt}
  \item \textbf{Relation Renaming}, which contains the same entity pair with the original one, but an alias relation name for generating a sentence with different expressions. Then this instance is considered as a positive sample to prototypical relation instance.
  \item \textbf{Context Expansion}, which extends the original relation instance with an additional triplet. The added triplet owns the same head/tail entity with the original instance but differs in the relation and tail/head entity. This variety aims to add irrelative context, which forces the model to focus on the important part of the context and is also considered as a positive sample to prototypical relation instance.
  \item \textbf{Relation Replacing}, which contains the same entity pair as the original one, but with other relations between these two entities. This variety aims to avoid spurious correlations that extracts only based on the entity pair and is considered as a negative instance to the prototypical relation instance.
\end{itemize}
Then we use the generator to generate texts based on these triplets.
Specifically, we first wrap the triplets with special markers \textit{``[H], [T],[ R]"} corresponds to head entity, tail entity, and relation name. Then we input the concatenated texts for relation instance generation. In our implementation, we use T5~\citep{raffel-2020, ribeiro2020} as the base generator, and pre-train the generator on WebNLG data~\citep{webnlg}. After sampled these intervened instances, we approximately optimize $\mathcal{P}(Y,P| do(E)) $ using the following contrastive loss function:
\begin{equation}
  % \begin{aligned}
  \setlength{\abovedisplayskip}{3pt}
  \setlength{\belowdisplayskip}{3pt}
  \begin{aligned}
    \mathcal{L}_{C}(\theta;\mathcal{X})=\sum_{X_p\in \mathcal{P}}\sum_{X_n\in \mathcal{N}}&\text{max}(D(P,X_p)\\
    -&D(P,X_n)+m_C,0),
  \end{aligned}
    % &\text{log}\sum_{x_i\in \mathcal{P}}e^{D(x_a,x_i)}\\
    % +&\text{log}\sum_{x_j\in \mathcal{N}}e^{m_C-D(x_a,x_j)}\\
  % \end{aligned}
\end{equation}
where $\theta$ is the model parameters, $\mathcal{X}$ is the intervened instance set, $\mathcal{P}$ is the positive instance set generated from relation renaming and context expansion, $\mathcal{N}$ is the negative instance set generated from relation replacing, $P$ is the original prototypical relation instance, $m_C$ is the margin.

\subsection{Surrogate Loss for Optimizing Causal Effects}
Based on entity ranking and context contrasting, we approximate the causal effects optimized in Equation (\ref{eq:loss}) with the following ranking and contrastive loss:
\begin{equation}
  \setlength{\abovedisplayskip}{3pt}
  \setlength{\belowdisplayskip}{3pt}
  \mathcal{L}(\theta;\mathcal{X})=\mathcal{L}_{E}(\theta;\mathcal{X})+\mathcal{L}_{C}(\theta;\mathcal{X}).
\end{equation}
which involves both the entity ranking loss and the context contrastive loss. During inference, we first encode each instance into its representation using the learned model. Then we apply a clustering algorithm to cluster the relation representations, and the relation for each instance is predicted through the clustering results.

% And we do not use any other kind of supervision from the OpenRE datasets during the training procedure.

% For relation prediction, we encode each instance in the unlabelled relation dataset to the representation space. 
% Based on these representations, we apply a clustering algorithm to cluster these instances, then get the cluster as the final relation prediction.

\begin{table*}[!tbp]
  \setlength{\belowcaptionskip}{-0.4cm}
  \centering
  \resizebox{0.9\textwidth}{!}{
    \begin{tabular}{|c|l|ccc|ccc|c|}
      \hline
      \multirow{2}[3]{*}{Dataset} &  \multirow{2}[3]{*}{model} & \multicolumn{3}{c|}{$\rm B^3$} & \multicolumn{3}{c|}{V-measure} &  \multirow{2}[3]{*}{ARI} \\
  \cline{3-8}          &       & F1 & Prec. & \multicolumn{1}{c|}{Rec.} & F1 & Homo. & \multicolumn{1}{c|}{Comp.} &  \\
      \hline
  %     \multirow{6}[18]{*}{NYT-FB} & rel-LDA-full & 36.9 & 30.4 & 47.0 & 37.4 & 31.9 & 45.1 & 24.2 \\
  % \cline{2-9}          & March & 35.2 & 23.8 & 67.1 & 27.0 &  18.6 &  49.6 &  18.7 \\
  % \cline{2-9}          & EType+ &  41.9 &  --  &  --  &  40.6 &  --   &   --  &  30.7 \\
  % \cline{2-9}          & UIE-PCNN &  39.4 &  32.2 &  50.7 &  38.3 &  32.2 &  47.2 &  33.8 \\
  % \cline{2-9}          & UIE-BERT &  41.5 &  34.6 &  51.8 &  39.9 &  33.9 &  48.5 &  35.1 \\
  % \cline{2-9}          & SelfORE &  49.1 &  47.3 &  51.1 &  46.6 &  45.7 &  47.6 &  40.3 \\
  % \cline{2-9}          & Our   &       &       &       &       &       &       &  \\
  % \cline{2-9}          & \ \ w/o context intvn. &       &       &       &       &       &       &  \\
  % \cline{2-9}          & \ \ w/o entity intvn. &       &       &       &       &       &       &  \\
  %     \hline
  %     \hline
      \multirow{5}[16]{*}{T-REx SPO} & rel-LDA-full~\citep{yao-2011}$^*$ &  18.5 &  14.3 &  26.1 &  19.4 &  16.1 &  24.5 &  8.6 \\
  \cline{2-9}          & March~\citep{marcheggiani-2016}$^*$ &  24.8 &  20.6 &  31.3 &  23.6 &  19.1 &  30.6 &  12.6 \\
  \cline{2-9}          & UIE-PCNN~\citep{simon-2019} &  36.3 &  28.4 &  50.3 &  41.4 &  33.7 &  53.6 &  21.3 \\
  \cline{2-9}          & UIE-BERT~\citep{simon-2019} &  38.1 &  30.7 &  50.3 &  39.1 &  37.6 &  40.8 &  23.5 \\
  \cline{2-9}          & SelfORE~\citep{hu-2020-selfore} &  41.0 &  39.4 &  42.8 &  41.4 &  40.3 &  42.5 &  33.7 \\
  \cline{2-9}          & Our   &   \textbf{45.0}   &  46.7   &  43.4   &    \textbf{45.3}  &   45.4   &  45.2 & \textbf{36.6} \\
  \cline{2-9}          & \ \ w/o Hyber &    41.4   &   40.9   &   42.0    &   43.7    &    42.3  &   45.2  & 33.2 \\
  \cline{2-9}          & \ \ w/o Gcc &   42.2    &  44.2   &  40.4   &   45.2   &   44.7  &    45.7   & 34.7 \\
      \hline
      \hline
      \multirow{5}[15]{*}{T-REx DS} & rel-LDA-full~\citep{yao-2011}$^*$ &  12.7 &  8.3 &  26.6 &  17.0 &  13.3 &  23.5 &  3.4 \\
  \cline{2-9}          & March~\citep{marcheggiani-2016}$^*$ &  9.0 &  6.4 &  15.5 &  5.7 &  4.5 &  7.9 &  1.9 \\
  \cline{2-9}          & UIE-PCNN~\citep{simon-2019} &  19.7 &  14.0 &  33.4 &  26.6 &  20.8 &  36.8 &  9.4 \\
  \cline{2-9}          & UIE-BERT~\citep{simon-2019} &  22.4 &  17.6 &  30.8 &  31.2 &  26.3 &  38.3 &  12.3 \\
  \cline{2-9}          & SelfORE~\citep{hu-2020-selfore} &  32.9 &  29.7 &  36.8 &  32.4 &  30.1 &  35.1 &  20.1 \\
  \cline{2-9}          & Our   & \textbf{42.9} &40.2 & 45.9 & \textbf{47.3} &46.9 & 47.8 &  \textbf{25.0}  \\
  \cline{2-9}          & \ \ w/o Hyber &  40.9 & 39.2 & 42.7  & 43.0 & 42.5 & 43.6 & 22.4   \\
  \cline{2-9}          & \ \ w/o Gcc & 41.5 & 40.1 & 42.9 & 45.2 & 44.8 & 45.6 & 21.7 \\
  \hline
      \end{tabular}%
    \label{tab:main}%
  }
  \caption{Results~(\%) on unsupervised relation extraction datasets. The results of * are reproduced in \citet{simon-2019}, Hyber refers to our Hierarchy-based Entity Ranking methods and Gcc refers to Generation-based Context Contrasting method.}
  \label{exp:main}%
\end{table*}%

\section{Experiments}
\subsection{Dataset}
We conduct experiments on two OpenRE datasets -- T-REx SPO and T-REx DS, since these datasets are from the same data source but only differ in constructing settings, which is very suitable for evaluating the stability of OpenRE methods. These datasets are both from T-REx\footnote{\href{https://hadyelsahar.github.io/t-rex/}{https://hadyelsahar.github.io/t-rex/}}~\citep{elsahar-2018} -- a dataset consists of Wikipedia sentences that are distantly aligned with Wikidata relation triplets; and these aligned sentences are further collected as T-REx SPO and T-REx DS according to whether they have surface-form relations or not.  As a result, T-REx SPO contains 763,000 sentences of 615 relations, and T-REx DS contains nearly 12 million sentences of 1189 relations. For both datasets, we use 20\% for validation and the remaining for model training as \citet{hu-2020-selfore}.

\subsection{Baseline and Evaluation Metrics}
\paragraph{Baseline Methods.}
We compare our model with the following baselines: 1)~\textbf{rel-LDA}~\citep{yao-2011}, a generative model that considers the unsupervised relation extraction as a topic model. We choose the full rel-LDA with a total number of 8 features for comparison in our experiment. 2)~\textbf{March}~\citep{marcheggiani-2016}, a VAE-based model learned by self-supervised signal of entity link predictor. 3)~\textbf{UIE}~\citep{simon-2019}, a discriminative model that adopts additional regularization to guide model learning. And it has different versions according to the choices of different relation encoding models (e.g., PCNN). We report the results of two versions—UIE-PCNN and UIE-BERT (i.e., using PCNN and BERT as the relation encoding models) with the highest performance. 4)~\textbf{SelfORE}~\citep{hu-2020-selfore}, a self-supervised framework that bootstraps to learn a contextual relation representation through adaptive clustering and pseudo label.

\paragraph{Evaluation Metrics.}
We adopt three commonly-used metrics to evaluate different methods: ${\rm B}^3$~\citep{b3-1998}, V-measure~\citep{v-measure-2007} and Adjusted Rand Index~(ARI)~\citep{ari-1985}.

Specifically,
${\rm B}^3$ 
contains the precision and recall metrics to correspondingly measure the correct rate of putting each sentence in its cluster or clustering all samples into a single class, which are defined as follows:
\begin{equation*}
  \setlength{\abovedisplayskip}{3pt}
  \setlength{\belowdisplayskip}{3pt}
  \begin{aligned}
    \rm B^3_{Prec.}=& \mathop{\mathbb{E}}\limits_{X,Y}P(g(X)=g(Y)|c(X)=c(Y))\\
    \rm B^3_{Rec.}=& \mathop{\mathbb{E}}\limits_{X,Y}P(c(X)=c(Y)|g(X)=g(Y))\\
  \end{aligned}
\end{equation*}
Then ${\rm B}^3$\ \ $\rm F_1$ is computed as the harmonic mean of the precision and recall.

Similar  to $\rm B^3$, V-measure focuses more on small impurities in a relatively ``pure" cluster than less ``pure" cluster, and use the homogeneity and completeness metrics:
\begin{equation*}
  \setlength{\abovedisplayskip}{3pt}
  \setlength{\belowdisplayskip}{3pt}
  \begin{aligned}
    \rm V_{Homo.}=& 1- H(c(X)|g(X))/H(c(X))\\
    \rm V_{Comp.}=& 1-H(g(X)|c(X))/H(g(x))\\
  \end{aligned}
\end{equation*}

ARI is a normalization of the Rand Index, which measures the agreement degree between the cluster and golden distribution. This metric ranges in [-1,1], a more accurate cluster will get a higher score.
Different from previous metrics, ARI is less sensitive to precision/homogeneity and recall/completeness.

\subsection{Hyperparameters and Implementation Details}
In the training period, we manually search the Hyperparameters of learning rate in [5e-6,1e-5, 5e-5], and find 1e-5 is optimal, search weight decay in [1e-6, 3e-6, 5e-5]  and choose 3e-6, and use other hyperparameters without search: the dropout rate of 0.6, the batch size of 32, and a linear learning schedule with a 0.85 decay rate per 1000 mini-batches.
In the evaluation period, we simply adopt the pre-trained models for representation extraction, then cluster the evaluation instances based on these representations. For clustering, we follow previous work~\citep{simon-2019,hu-2020-selfore} and set $K$=10 as the number of clusters. The training period of each epoch costs about one day.
In our implementation, we adopt Bert-base-uncased model~\footnote{\href{https://github.com/huggingface/transformers}{https://github.com/huggingface/transformers}} as the base model for relation extraction and a modified T5-base model~\footnote{\href{https://github.com/UKPLab/plms-graph2text}{https://github.com/UKPLab/plms-graph2text}} for text generation.
The entity hierarchical tree is constructed based on WikiData and finally contains 589,121 entities. The generation set contains about 530,000 triplets, and each triplet corresponds to 5 positive/negative triplets and generated texts. 
We use one Titan RTX for Element Intervention training and four cards of RTX for text generation.

\begin{table}[!tbp]
  \setlength{\belowcaptionskip}{-0.4cm}
  \centering
    \begin{tabular}{c|ccc}
    Source  & $\rm B^3$ & V-meas.  & ARI \\
    \hline
    \hline
    T-REx SPO &  45.0   &  45.3  &  36.6 \\
    Generated &   46.0   &  44.6  &  36.7 
    \end{tabular}%
    \caption{The results~(\%) of entity ranking based on different data sources. These results are reported on T-REx SPO. And we only report the $\rm F_1$ scores of $\rm B^3$ and V-measure for simplicity.}
  \label{tab:ent}%
\end{table}%

\subsection{Overall Results}
Table~\ref{exp:main} shows the overall results on T-REx SPO and T-REx DS.
From this table, we can see that:
\begin{enumerate}
  \item \textbf{Our method outperforms previous OpenRE models and achieves the new state-of-the-art performance.} Comparing with all baseline models, our method achieves significant performance improvements: on T-Rex SPO, our method improves the SOTA $\rm B^3\ F_1$ and V-measure $\rm F_1$ by at least 3.9\%, and ARI by 2.9\%; on T-Rex DS, the improvements are more evident, where SOTA $\rm B^3\ F_1$  and V-measure $\rm F_1$ are improved by at least 10.0\%, and ARI is improved by 4.9\%.
  \item \textbf{Our method performs robustly in different datasets.} Comparing the performances on these two datasets, we can see that almost all baseline methods suffer dramatic performance drops on all these metrics, which verifies that previous OpenRE methods can be easily influenced by the spurious correlations in datasets, as T-REx DS involves much more noisy instances without relation surface forms. As contrast, our method has marginal performance differences, which indicates both the effectiveness and robustness of our method.
\end{enumerate}

\subsection{Detailed Analysis}
In this section, we conduct several experiments for detailed analysis of our method. 

\paragraph{Ablation Study.}
To study the effect of different intervention modules, we conduct an ablation study on each intervention module by correspondingly ablating one. The other setting remains the same as the main model. From Table~\ref{exp:main}, we can see that,
in both T-REx SPO and DS, combining these two modules can result in a noticeable performance gain, which demonstrates that both two modules are important to the final model performance and they are complementary on alleviating unnecessary co-dependencies: Hyber aims to alleviate the spurious correlations between the context and the final relation prediction, and Gcc aims to alleviate the spurious correlations between entity pair and the final relation prediction.
Besides, in T-REx DS, we can see that Hyber or Gcc only is effective enough to outperform previous SOTA methods, which indicates that element intervention has clearly unbiased representation on either entity pair or context.

\paragraph{Entity Ranking on Generated Texts.}
This experiment studies the effect of different data sources for Hyber module.
As shown in Table~\ref{tab:ent}, we can see that Hyber based on T-REx SPO dataset or the generated texts has marginal difference. 
That means Hyber is robust to the source context. On the other hand, the quality of the generated texts satisfies the demand of this task.

\begin{table}[!tbp]
  \setlength{\belowcaptionskip}{-0.4cm}
  \centering
    \begin{tabular}{c|ccc}
    Metrics & Both  & Seen  & Unseen \\
    \hline
    \hline
    BLEU  & 60.9 & 65.9 & 54.9 \\
    chrF++ & 76.0 & 79.2 & 72.5 \\
    \end{tabular}%
    \caption{Quantitative performance of our generator on WebNLG. Seen stands for generating from seen relation triplets, unseen stands for generating from unseen relation triplets. Both stands for a combination of seen and unseen relation triplets.}
  \label{tab:gen}%
\end{table}%
\setlength{\parskip}{0em}
\paragraph{Quality of Context Generation(unseen relations).}
This experiment gives a quantitative analysis of the generator used in our work. 
We select WebNLG~\citep{webnlg} to test the generator, and adopt the widely-used metrics including BLEU~\citep{papineni-bleu} and chrF++~\citep{popovic-2017-chrf} for evaluation. 
As shown in Table~\ref{tab:gen}, we can see that our generator is quite effective on seen relation generation.
Though the generator suffers a performance drop in unseen relations, the scores are still receptible. 
Combined with results from other experiments, the generator is sufficient for this task.

\begin{figure}[!tpb] 
  \setlength{\belowcaptionskip}{-0.4cm}
  \centering
  \includegraphics[width=0.35\textwidth]{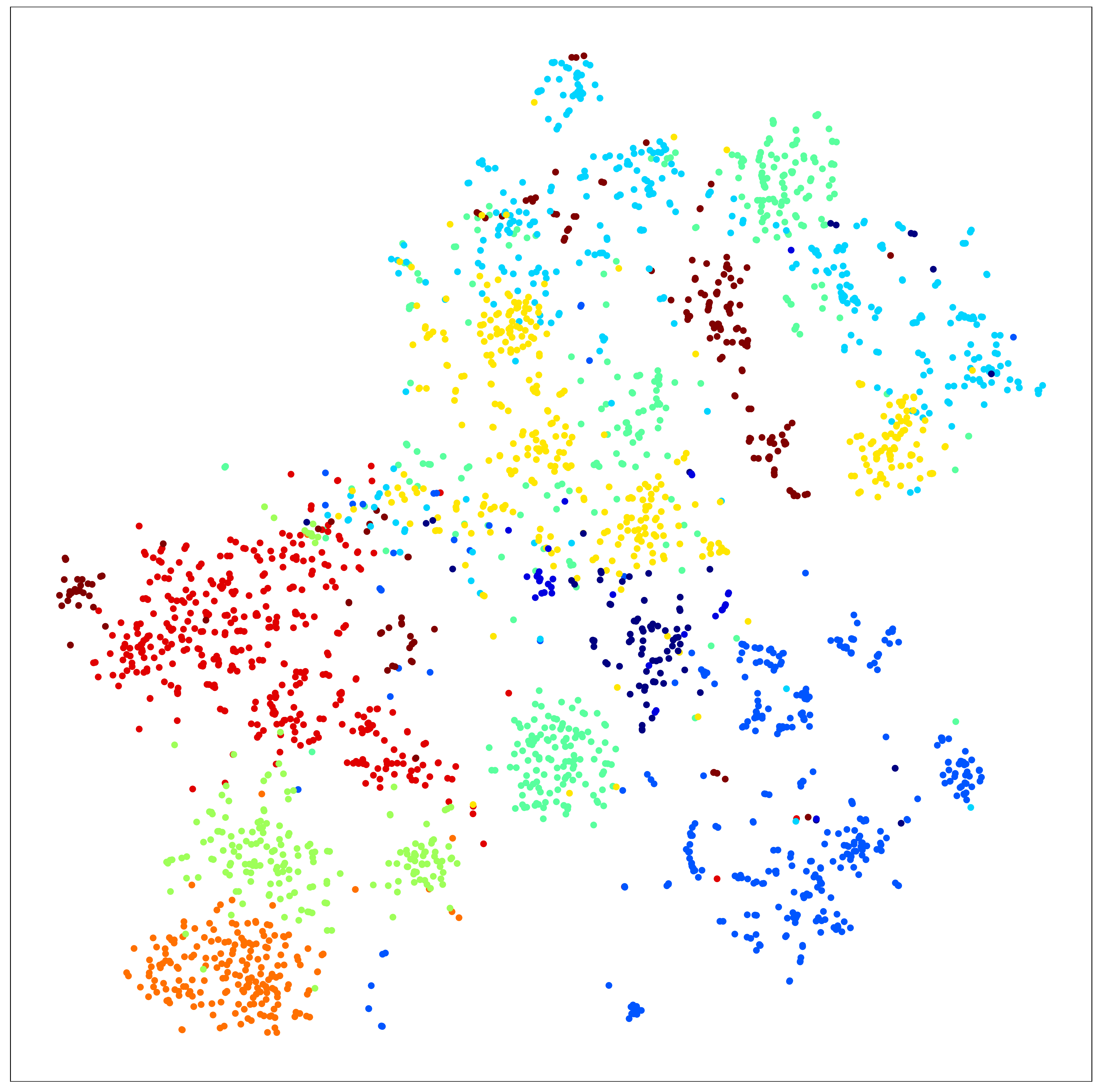}
    \caption{Visualization of relation representation learned by element intervention. Each relation instance is colored with the ground-truth label.}
  \label{fig:vis}
\end{figure}

\setlength{\parskip}{0em}
\paragraph{Visualization of Relation Representations.}
In this experiment, we visual the representations of the validation instances. 
We sample 10 relations from the T-REx SPO validation set and each relation with 200 instances for visualization.
To reduce the dimension, we use t-sne~\citep{t-sne} to map each representation to the dimension of 2.
For the convenience of comparison, we color each instance with its ground-truth relation label.
Since the visualization results of only Hyber or Gcc are marginally different from the full model, so we only choose the full model for visualization.
As shown in Figure~\ref{fig:vis}, we can see that each relation is mostly separate from others. However, there still be some instances misclassified due to the overlapping in the representation space.

\setlength{\parskip}{0em}
\section{Related Work}
Current success of supervised relation extraction methods~\citep{bunescu-2005,qian-2008,zeng-2014,zhou-2016,Velickovic-2018} depends heavily on large amount of annotated data. Due to this data bottleneck, some weakly-supervised methods are proposed to learn relation extraction models from distantly labeled datasets~\citep{mintz-2009,hoffmann-2011,lin-2016} or few-shot datasets~\citep{han-fewrel,soares-2019,peng-2020}.
However, these paradigms still require pre-defined relation types and therefore restricts their application to open scenarios.

Open relation extraction, on the other hand, aims to cluster relation instances referring to the same underlying relation without pre-defined relation types. Previous methods for OpenRE can be roughly divided into two categories.
The generative method~\citep{yao-2011} formulates OpenRE using a topic model, and the latent relations are generated based on the hand-crafted feature representations of entities and context. While the discriminative method is first proposed by ~\citet{marcheggiani-2016}, which learns the model through the self-supervised signal from entity link predictor. Along this line, \citet{hu-2020-selfore} propose the SelfORE that learns the model through pseudo label and bootstrapping technology. However, \citet{simon-2019} point out that previous OpenRE methods severely suffer from the instability, and they also propose two regularizers to guide the learning procedure. But the fundamental cause of the instability is still undiscovered.

In this paper, we revisit the procedure of OpenRE from a causal view. By formulating OpenRE using a structural causal model, we identify the cause of the above-mentioned problems, and alleviate the problems by Element Intervention. There are also some recent studies try to introduce causal theory to explain the spurious correlations in neural models~\citep{feng-2018,gururangan-2018,tang2020longtailed,Qi_2020_CVPR,zeng-2020,wu-2020,qin-2020,fu-2020}. However, to the best of our knowledge, this is the first work to revisit OpenRE from the perspective of causality.

\setlength{\parskip}{0em}
\section{Conclusions}
In this paper, we revisit OpenRE from the perspective of causal theory. We find that the strong connections between the generated instance to the prototypical instance through either their entities or their context will result in spurious correlations, which appear in the form of the backdoor paths in the SCM. Then the spurious correlations will mislead OpenRE models. Based on the observations, we propose \emph{Element Intervention} to block the backdoor paths, which intervenes on the context and entities respectively to obtain the underlying causal effects of them. We also provide two specific implementations of the interventions based on entity ranking and context contrasting.
Experimental results on two OpenRE datasets show that our methods outperform previous methods with a large margin, and suffer the least performance discrepancy between datasets, which indicates both the effectiveness and stability of our methods.

\section*{Acknowledgements}
We thank all reviewers for their insightful suggestions. Moreover, this work is supported by the National Key Research and Development Program of China under Grant No.2019YFC1521200, the National Natural Science Foundation of China under Grants no. U1936207 and 61772505, and in part by the Youth Innovation Promotion Association CAS(2018141).

\bibliographystyle{acl_natbib}
\bibliography{ca_re}

\begin{thebibliography}{39}
\expandafter\ifx\csname natexlab\endcsname\relax\def\natexlab#1{#1}\fi

\bibitem[{Bagga and Baldwin(1998)}]{b3-1998}
Amit Bagga and Breck Baldwin. 1998.
\newblock \href {https://doi.org/10.3115/980845.980859} {Entity-based
  cross-document coreferencing using the vector space model}.
\newblock In \emph{36th Annual Meeting of the Association for Computational
  Linguistics and 17th International Conference on Computational Linguistics,
  Volume 1}, pages 79--85, Montreal, Quebec, Canada. Association for
  Computational Linguistics.

\bibitem[{Baldini~Soares et~al.(2019)Baldini~Soares, FitzGerald, Ling, and
  Kwiatkowski}]{soares-2019}
Livio Baldini~Soares, Nicholas FitzGerald, Jeffrey Ling, and Tom Kwiatkowski.
  2019.
\newblock \href {https://doi.org/10.18653/v1/P19-1279} {Matching the blanks:
  Distributional similarity for relation learning}.
\newblock In \emph{Proceedings of the 57th Annual Meeting of the Association
  for Computational Linguistics}, pages 2895--2905, Florence, Italy.
  Association for Computational Linguistics.

\bibitem[{Banko et~al.(2007)Banko, Cafarella, Soderland, Broadhead, and
  Etzioni}]{banko-2007}
Michele Banko, Michael~J. Cafarella, Stephen Soderland, Matthew Broadhead, and
  Oren Etzioni. 2007.
\newblock \href {http://ijcai.org/Proceedings/07/Papers/429.pdf} {Open
  information extraction from the web}.
\newblock In \emph{{IJCAI} 2007, Proceedings of the 20th International Joint
  Conference on Artificial Intelligence, Hyderabad, India, January 6-12, 2007},
  pages 2670--2676.

\bibitem[{Blei et~al.(2003)Blei, Ng, and Jordan}]{blei-lda}
David~M. Blei, Andrew~Y. Ng, and Michael~I. Jordan. 2003.
\newblock \href {https://doi.org/http://dx.doi.org/10.1162/jmlr.2003.3.4-5.993}
  {Latent dirichlet allocation}.
\newblock \emph{Journal of Machine Learning Research}, 3:993--1022.

\bibitem[{Bunescu and Mooney(2005)}]{bunescu-2005}
Razvan Bunescu and Raymond Mooney. 2005.
\newblock \href {https://www.aclweb.org/anthology/H05-1091} {A shortest path
  dependency kernel for relation extraction}.
\newblock In \emph{Proceedings of Human Language Technology Conference and
  Conference on Empirical Methods in Natural Language Processing}, pages
  724--731, Vancouver, British Columbia, Canada. Association for Computational
  Linguistics.

\bibitem[{Devlin et~al.(2019)Devlin, Chang, Lee, and Toutanova}]{devlin-2019}
Jacob Devlin, Ming-Wei Chang, Kenton Lee, and Kristina Toutanova. 2019.
\newblock \href {https://doi.org/10.18653/v1/N19-1423} {{BERT}: Pre-training of
  deep bidirectional transformers for language understanding}.
\newblock In \emph{Proceedings of the 2019 Conference of the North {A}merican
  Chapter of the Association for Computational Linguistics: Human Language
  Technologies, Volume 1 (Long and Short Papers)}, pages 4171--4186,
  Minneapolis, Minnesota. Association for Computational Linguistics.

\bibitem[{Elsahar et~al.(2018)Elsahar, Vougiouklis, Remaci, Gravier, Hare,
  Laforest, and Simperl}]{elsahar-2018}
Hady Elsahar, Pavlos Vougiouklis, Arslen Remaci, Christophe Gravier, Jonathon
  Hare, Frederique Laforest, and Elena Simperl. 2018.
\newblock \href {https://www.aclweb.org/anthology/L18-1544} {{T}-{RE}x: A large
  scale alignment of natural language with knowledge base triples}.
\newblock In \emph{Proceedings of the Eleventh International Conference on
  Language Resources and Evaluation ({LREC} 2018)}, Miyazaki, Japan. European
  Language Resources Association (ELRA).

\bibitem[{Feng et~al.(2018)Feng, Wallace, Grissom~II, Iyyer, Rodriguez, and
  Boyd-Graber}]{feng-2018}
Shi Feng, Eric Wallace, Alvin Grissom~II, Mohit Iyyer, Pedro Rodriguez, and
  Jordan Boyd-Graber. 2018.
\newblock \href {https://doi.org/10.18653/v1/D18-1407} {Pathologies of neural
  models make interpretations difficult}.
\newblock In \emph{Proceedings of the 2018 Conference on Empirical Methods in
  Natural Language Processing}, pages 3719--3728, Brussels, Belgium.
  Association for Computational Linguistics.

\bibitem[{Fu et~al.(2020)Fu, Wang, Grafton, Eckstein, and Wang}]{fu-2020}
Tsu-Jui Fu, Xin Wang, Scott Grafton, Miguel Eckstein, and William~Yang Wang.
  2020.
\newblock \href {https://doi.org/10.18653/v1/2020.emnlp-main.357} {{SSCR}:
  Iterative language-based image editing via self-supervised counterfactual
  reasoning}.
\newblock In \emph{Proceedings of the 2020 Conference on Empirical Methods in
  Natural Language Processing (EMNLP)}, pages 4413--4422, Online. Association
  for Computational Linguistics.

\bibitem[{Gardent et~al.(2017)Gardent, Shimorina, Narayan, and
  Perez-Beltrachini}]{webnlg}
Claire Gardent, Anastasia Shimorina, Shashi Narayan, and Laura
  Perez-Beltrachini. 2017.
\newblock \href {https://doi.org/10.18653/v1/W17-3518} {The {W}eb{NLG}
  challenge: Generating text from {RDF} data}.
\newblock In \emph{Proceedings of the 10th International Conference on Natural
  Language Generation}, pages 124--133, Santiago de Compostela, Spain.
  Association for Computational Linguistics.

\bibitem[{Gururangan et~al.(2018)Gururangan, Swayamdipta, Levy, Schwartz,
  Bowman, and Smith}]{gururangan-2018}
Suchin Gururangan, Swabha Swayamdipta, Omer Levy, Roy Schwartz, Samuel Bowman,
  and Noah~A. Smith. 2018.
\newblock \href {https://doi.org/10.18653/v1/N18-2017} {Annotation artifacts in
  natural language inference data}.
\newblock In \emph{Proceedings of the 2018 Conference of the North {A}merican
  Chapter of the Association for Computational Linguistics: Human Language
  Technologies, Volume 2 (Short Papers)}, pages 107--112, New Orleans,
  Louisiana. Association for Computational Linguistics.

\bibitem[{Han et~al.(2018)Han, Zhu, Yu, Wang, Yao, Liu, and Sun}]{han-fewrel}
Xu~Han, Hao Zhu, Pengfei Yu, Ziyun Wang, Yuan Yao, Zhiyuan Liu, and Maosong
  Sun. 2018.
\newblock \href {https://doi.org/10.18653/v1/D18-1514} {{F}ew{R}el: A
  large-scale supervised few-shot relation classification dataset with
  state-of-the-art evaluation}.
\newblock In \emph{Proceedings of the 2018 Conference on Empirical Methods in
  Natural Language Processing}, pages 4803--4809, Brussels, Belgium.
  Association for Computational Linguistics.

\bibitem[{Hoffmann et~al.(2011)Hoffmann, Zhang, Ling, Zettlemoyer, and
  Weld}]{hoffmann-2011}
Raphael Hoffmann, Congle Zhang, Xiao Ling, Luke Zettlemoyer, and Daniel~S.
  Weld. 2011.
\newblock \href {https://www.aclweb.org/anthology/P11-1055} {Knowledge-based
  weak supervision for information extraction of overlapping relations}.
\newblock In \emph{Proceedings of the 49th Annual Meeting of the Association
  for Computational Linguistics: Human Language Technologies}, pages 541--550,
  Portland, Oregon, USA. Association for Computational Linguistics.

\bibitem[{Hu et~al.(2020)Hu, Wen, Xu, Zhang, and Yu}]{hu-2020-selfore}
Xuming Hu, Lijie Wen, Yusong Xu, Chenwei Zhang, and Philip Yu. 2020.
\newblock \href {https://doi.org/10.18653/v1/2020.emnlp-main.299} {{S}elf{ORE}:
  Self-supervised relational feature learning for open relation extraction}.
\newblock In \emph{Proceedings of the 2020 Conference on Empirical Methods in
  Natural Language Processing (EMNLP)}, pages 3673--3682, Online. Association
  for Computational Linguistics.

\bibitem[{Hubert and Arabie(1985)}]{ari-1985}
L.~Hubert and P.~Arabie. 1985.
\newblock \href
  {http://scholar.google.de/scholar.bib?q=info:IkrWWF2JxwoJ:scholar.google.com/&output=citation&hl=de&ct=citation&cd=0}
  {{Comparing partitions}}.
\newblock \emph{Journal of classification}, 2(1):193--218.

\bibitem[{Lewis et~al.(2020)Lewis, Liu, Goyal, Ghazvininejad, Mohamed, Levy,
  Stoyanov, and Zettlemoyer}]{bart}
Mike Lewis, Yinhan Liu, Naman Goyal, Marjan Ghazvininejad, Abdelrahman Mohamed,
  Omer Levy, Veselin Stoyanov, and Luke Zettlemoyer. 2020.
\newblock \href {https://doi.org/10.18653/v1/2020.acl-main.703} {{BART}:
  Denoising sequence-to-sequence pre-training for natural language generation,
  translation, and comprehension}.
\newblock In \emph{Proceedings of the 58th Annual Meeting of the Association
  for Computational Linguistics}, pages 7871--7880, Online. Association for
  Computational Linguistics.

\bibitem[{Lin et~al.(2016)Lin, Shen, Liu, Luan, and Sun}]{lin-2016}
Yankai Lin, Shiqi Shen, Zhiyuan Liu, Huanbo Luan, and Maosong Sun. 2016.
\newblock \href {https://doi.org/10.18653/v1/P16-1200} {Neural relation
  extraction with selective attention over instances}.
\newblock In \emph{Proceedings of the 54th Annual Meeting of the Association
  for Computational Linguistics (Volume 1: Long Papers)}, pages 2124--2133,
  Berlin, Germany. Association for Computational Linguistics.

\bibitem[{van~der Maaten and Hinton(2008)}]{t-sne}
Laurens van~der Maaten and Geoffrey Hinton. 2008.
\newblock \href {http://www.jmlr.org/papers/v9/vandermaaten08a.html}
  {Visualizing data using {t-SNE}}.
\newblock \emph{Journal of Machine Learning Research}, 9:2579--2605.

\bibitem[{Marcheggiani and Titov(2016)}]{marcheggiani-2016}
Diego Marcheggiani and Ivan Titov. 2016.
\newblock \href {https://doi.org/10.1162/tacl_a_00095} {Discrete-state
  variational autoencoders for joint discovery and factorization of relations}.
\newblock \emph{Transactions of the Association for Computational Linguistics},
  4:231--244.

\bibitem[{Mintz et~al.(2009)Mintz, Bills, Snow, and Jurafsky}]{mintz-2009}
Mike Mintz, Steven Bills, Rion Snow, and Daniel Jurafsky. 2009.
\newblock \href {https://www.aclweb.org/anthology/P09-1113} {Distant
  supervision for relation extraction without labeled data}.
\newblock In \emph{Proceedings of the Joint Conference of the 47th Annual
  Meeting of the {ACL} and the 4th International Joint Conference on Natural
  Language Processing of the {AFNLP}}, pages 1003--1011, Suntec, Singapore.
  Association for Computational Linguistics.

\bibitem[{Papineni et~al.(2002)Papineni, Roukos, Ward, and Zhu}]{papineni-bleu}
Kishore Papineni, Salim Roukos, Todd Ward, and Wei-Jing Zhu. 2002.
\newblock \href {https://doi.org/10.3115/1073083.1073135} {{B}leu: a method for
  automatic evaluation of machine translation}.
\newblock In \emph{Proceedings of the 40th Annual Meeting of the Association
  for Computational Linguistics}, pages 311--318, Philadelphia, Pennsylvania,
  USA. Association for Computational Linguistics.

\bibitem[{Pearl(2009)}]{pearl_2009}
Judea Pearl. 2009.
\newblock \emph{Causality: {Models}, Reasoning, and Inference}.
\newblock Cambridge University Press.

\bibitem[{Peng et~al.(2020)Peng, Gao, Han, Lin, Li, Liu, Sun, and
  Zhou}]{peng-2020}
Hao Peng, Tianyu Gao, Xu~Han, Yankai Lin, Peng Li, Zhiyuan Liu, Maosong Sun,
  and Jie Zhou. 2020.
\newblock \href {https://doi.org/10.18653/v1/2020.emnlp-main.298} {{L}earning
  from {C}ontext or {N}ames? {A}n {E}mpirical {S}tudy on {N}eural {R}elation
  {E}xtraction}.
\newblock In \emph{Proceedings of the 2020 Conference on Empirical Methods in
  Natural Language Processing (EMNLP)}, pages 3661--3672, Online. Association
  for Computational Linguistics.

\bibitem[{Popovi{\'c}(2017)}]{popovic-2017-chrf}
Maja Popovi{\'c}. 2017.
\newblock \href {https://doi.org/10.18653/v1/W17-4770} {chr{F}++: words helping
  character n-grams}.
\newblock In \emph{Proceedings of the Second Conference on Machine
  Translation}, pages 612--618, Copenhagen, Denmark. Association for
  Computational Linguistics.

\bibitem[{Qi et~al.(2020)Qi, Niu, Huang, and Zhang}]{Qi_2020_CVPR}
Jiaxin Qi, Yulei Niu, Jianqiang Huang, and Hanwang Zhang. 2020.
\newblock Two causal principles for improving visual dialog.
\newblock In \emph{Proceedings of the IEEE/CVF Conference on Computer Vision
  and Pattern Recognition (CVPR)}.

\bibitem[{Qian et~al.(2008)Qian, Zhou, Kong, Zhu, and Qian}]{qian-2008}
Longhua Qian, Guodong Zhou, Fang Kong, Qiaoming Zhu, and Peide Qian. 2008.
\newblock \href {https://www.aclweb.org/anthology/C08-1088} {Exploiting
  constituent dependencies for tree kernel-based semantic relation extraction}.
\newblock In \emph{Proceedings of the 22nd International Conference on
  Computational Linguistics (Coling 2008)}, pages 697--704, Manchester, UK.
  Coling 2008 Organizing Committee.

\bibitem[{Qin et~al.(2020)Qin, Shwartz, West, Bhagavatula, Hwang, Le~Bras,
  Bosselut, and Choi}]{qin-2020}
Lianhui Qin, Vered Shwartz, Peter West, Chandra Bhagavatula, Jena~D. Hwang,
  Ronan Le~Bras, Antoine Bosselut, and Yejin Choi. 2020.
\newblock \href {https://doi.org/10.18653/v1/2020.emnlp-main.58} {Back to the
  future: Unsupervised backprop-based decoding for counterfactual and abductive
  commonsense reasoning}.
\newblock In \emph{Proceedings of the 2020 Conference on Empirical Methods in
  Natural Language Processing (EMNLP)}, pages 794--805, Online. Association for
  Computational Linguistics.

\bibitem[{Radford et~al.(2019)Radford, Wu, Child, Luan, Amodei, and
  Sutskever}]{gpt2}
Alec Radford, Jeffrey Wu, Rewon Child, David Luan, Dario Amodei, and Ilya
  Sutskever. 2019.
\newblock Language models are unsupervised multitask learners.
\newblock \emph{OpenAI blog}, 1(8):9.

\bibitem[{Raffel et~al.(2020)Raffel, Shazeer, Roberts, Lee, Narang, Matena,
  Zhou, Li, and Liu}]{raffel-2020}
Colin Raffel, Noam Shazeer, Adam Roberts, Katherine Lee, Sharan Narang, Michael
  Matena, Yanqi Zhou, Wei Li, and Peter~J. Liu. 2020.
\newblock \href {http://jmlr.org/papers/v21/20-074.html} {Exploring the limits
  of transfer learning with a unified text-to-text transformer}.
\newblock \emph{Journal of Machine Learning Research}, 21(140):1--67.

\bibitem[{Ribeiro et~al.(2020)Ribeiro, Schmitt, Sch{\"{u}}tze, and
  Gurevych}]{ribeiro2020}
Leonardo F.~R. Ribeiro, Martin Schmitt, Hinrich Sch{\"{u}}tze, and Iryna
  Gurevych. 2020.
\newblock \href {http://arxiv.org/abs/2007.08426} {Investigating pretrained
  language models for graph-to-text generation}.
\newblock \emph{CoRR}, abs/2007.08426.

\bibitem[{Rosenberg and Hirschberg(2007)}]{v-measure-2007}
Andrew Rosenberg and Julia Hirschberg. 2007.
\newblock \href {https://www.aclweb.org/anthology/D07-1043} {{V}-measure: A
  conditional entropy-based external cluster evaluation measure}.
\newblock In \emph{Proceedings of the 2007 Joint Conference on Empirical
  Methods in Natural Language Processing and Computational Natural Language
  Learning ({EMNLP}-{C}o{NLL})}, pages 410--420, Prague, Czech Republic.
  Association for Computational Linguistics.

\bibitem[{Simon et~al.(2019)Simon, Guigue, and Piwowarski}]{simon-2019}
{\'E}tienne Simon, Vincent Guigue, and Benjamin Piwowarski. 2019.
\newblock \href {https://doi.org/10.18653/v1/P19-1133} {Unsupervised
  information extraction: Regularizing discriminative approaches with relation
  distribution losses}.
\newblock In \emph{Proceedings of the 57th Annual Meeting of the Association
  for Computational Linguistics}, pages 1378--1387, Florence, Italy.
  Association for Computational Linguistics.

\bibitem[{Tang et~al.(2020)Tang, Huang, and Zhang}]{tang2020longtailed}
Kaihua Tang, Jianqiang Huang, and Hanwang Zhang. 2020.
\newblock Long-tailed classification by keeping the good and removing the bad
  momentum causal effect.
\newblock In \emph{NeurIPS}.

\bibitem[{Veličković et~al.(2018)Veličković, Cucurull, Casanova, Romero,
  Liò, and Bengio}]{Velickovic-2018}
Petar Veličković, Guillem Cucurull, Arantxa Casanova, Adriana Romero, Pietro
  Liò, and Yoshua Bengio. 2018.
\newblock \href {https://openreview.net/forum?id=rJXMpikCZ} {Graph attention
  networks}.
\newblock In \emph{International Conference on Learning Representations}.

\bibitem[{Wu et~al.(2020)Wu, Kuang, Zhang, Liu, Sun, Xiao, Zhuang, Si, and
  Wu}]{wu-2020}
Yiquan Wu, Kun Kuang, Yating Zhang, Xiaozhong Liu, Changlong Sun, Jun Xiao,
  Yueting Zhuang, Luo Si, and Fei Wu. 2020.
\newblock \href {https://doi.org/10.18653/v1/2020.emnlp-main.56} {De-biased
  court{'}s view generation with causality}.
\newblock In \emph{Proceedings of the 2020 Conference on Empirical Methods in
  Natural Language Processing (EMNLP)}, pages 763--780, Online. Association for
  Computational Linguistics.

\bibitem[{Yao et~al.(2011)Yao, Haghighi, Riedel, and McCallum}]{yao-2011}
Limin Yao, Aria Haghighi, Sebastian Riedel, and Andrew McCallum. 2011.
\newblock \href {https://www.aclweb.org/anthology/D11-1135} {Structured
  relation discovery using generative models}.
\newblock In \emph{Proceedings of the 2011 Conference on Empirical Methods in
  Natural Language Processing}, pages 1456--1466, Edinburgh, Scotland, UK.
  Association for Computational Linguistics.

\bibitem[{Zeng et~al.(2014)Zeng, Liu, Lai, Zhou, and Zhao}]{zeng-2014}
Daojian Zeng, Kang Liu, Siwei Lai, Guangyou Zhou, and Jun Zhao. 2014.
\newblock \href {https://www.aclweb.org/anthology/C14-1220} {Relation
  classification via convolutional deep neural network}.
\newblock In \emph{Proceedings of {COLING} 2014, the 25th International
  Conference on Computational Linguistics: Technical Papers}, pages 2335--2344,
  Dublin, Ireland. Dublin City University and Association for Computational
  Linguistics.

\bibitem[{Zeng et~al.(2020)Zeng, Li, Zhai, and Zhang}]{zeng-2020}
Xiangji Zeng, Yunliang Li, Yuchen Zhai, and Yin Zhang. 2020.
\newblock \href {https://doi.org/10.18653/v1/2020.emnlp-main.590}
  {Counterfactual generator: A weakly-supervised method for named entity
  recognition}.
\newblock In \emph{Proceedings of the 2020 Conference on Empirical Methods in
  Natural Language Processing (EMNLP)}, pages 7270--7280, Online. Association
  for Computational Linguistics.

\bibitem[{Zhou et~al.(2016)Zhou, Shi, Tian, Qi, Li, Hao, and Xu}]{zhou-2016}
Peng Zhou, Wei Shi, Jun Tian, Zhenyu Qi, Bingchen Li, Hongwei Hao, and Bo~Xu.
  2016.
\newblock \href {https://doi.org/10.18653/v1/P16-2034} {Attention-based
  bidirectional long short-term memory networks for relation classification}.
\newblock In \emph{Proceedings of the 54th Annual Meeting of the Association
  for Computational Linguistics (Volume 2: Short Papers)}, pages 207--212,
  Berlin, Germany. Association for Computational Linguistics.

\end{thebibliography}

%\appendix

\end{document}